\documentclass[fullpaper,final]{nldl}

\paperID{58}

\vol{233}
\usepackage{url}
\usepackage{graphicx}
\usepackage{authblk}
\usepackage{bbm}
\usepackage{amsmath}
\usepackage{amsthm}
\newtheorem{theorem}{Theorem}
\usepackage{comment}

\usepackage{algorithm}
\usepackage{algorithmic}

\usepackage{lipsum}

\usepackage{hyperref}

\usepackage{bbm}
\usepackage{amsmath}
\usepackage{subcaption}

\title{
TraCE: Trajectory Counterfactual Explanation Scores
}
\author[1]{Jeffrey N. Clark\textsuperscript{\dag}\thanks{Corresponding Author: jeff.clark@bristol.ac.uk}}
\author[1,2]{Edward A. Small\thanks{Equal Contribution}}
\author[1]{Nawid Keshtmand}
\author[1]{Michelle W.L. Wan}
\author[1]{Elena Fillola Mayoral}
\author[1]{Enrico Werner}
\author[3]{Christopher P. Bourdeaux}
\author[1]{Raul Santos-Rodriguez}

\affil[1]{University of Bristol, UK}
\affil[2]{Royal Melbourne Institute of Technology, Australia}
\affil[3]{University Hospitals Bristol NHS Foundation Trust, UK}

\date{\vspace{-5ex}}

\begin{document}
\maketitle

\begin{abstract}
Counterfactual explanations, and their associated algorithmic recourse, are typically leveraged to understand and explain predictions of individual instances coming from a black-box classifier. In this paper, we propose to extend the use of counterfactuals to evaluate progress in sequential decision making tasks. To this end, we introduce a model-agnostic modular framework, TraCE (Trajectory Counterfactual Explanation) scores, to distill and condense progress in highly complex scenarios into a single value. We demonstrate TraCE's utility by showcasing its main properties in two case studies spanning healthcare and climate change.
\end{abstract}

\section{Introduction}
Counterfactual explanations can aid interpretation of predictions and address a lack of model transparency~\cite{wachter2018a}. 
For example, counterfactuals have been applied to the prediction of patient survival within an intensive care unit~\cite{wang2021counterfactual}. For an unwell patient predicted not to survive, a counterfactual and algorithmic recourse may demonstrate the feature changes necessary to result in positive survival classification. In this way counterfactuals aid users in understanding the model and may provide actionable input to support decisions.

\begin{figure*}[t]
    \centering
    \includegraphics[width=0.95\textwidth]{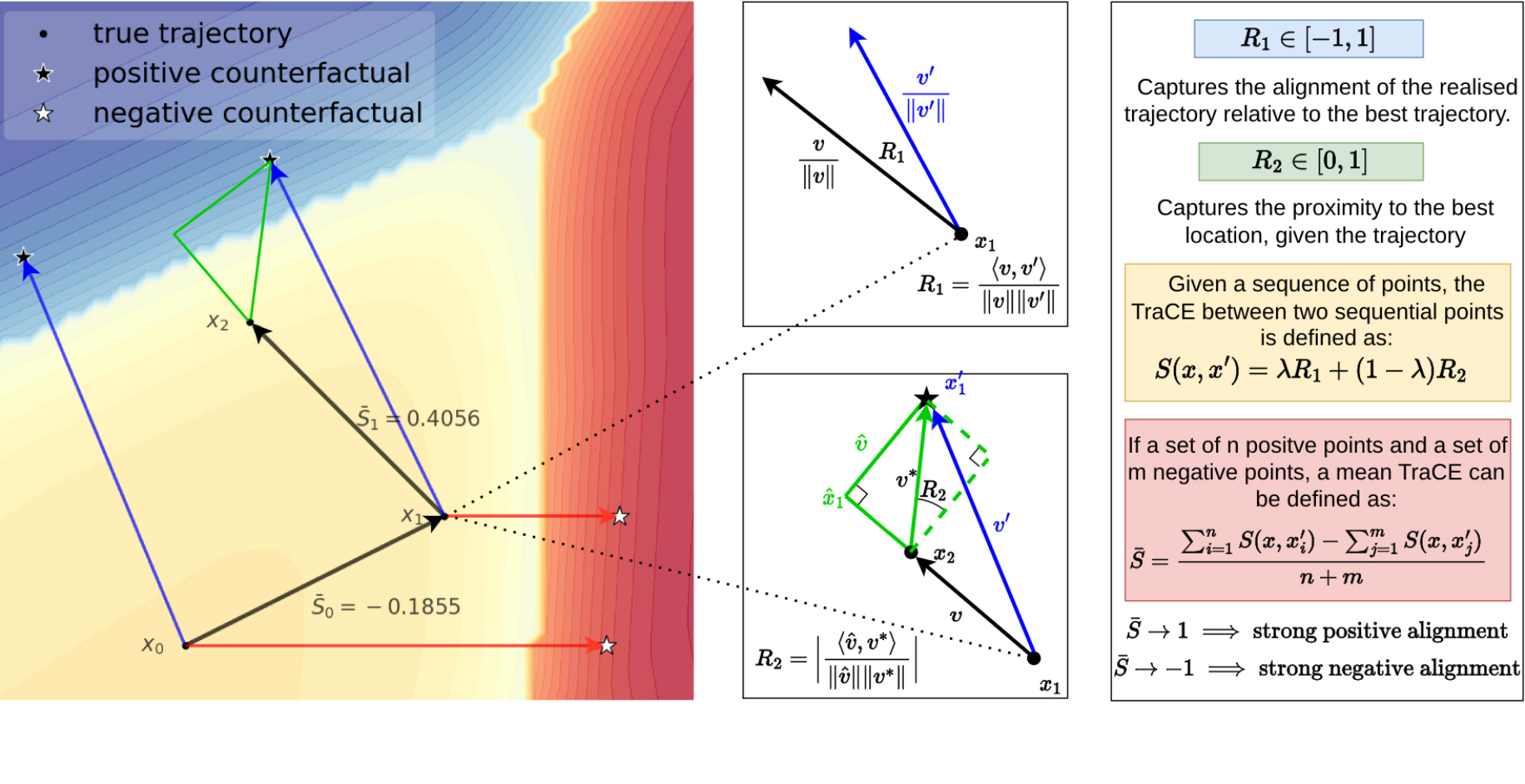}
    \caption{TraCE for 2-D toy data set classification with three classes: light orange (current class), blue (desired class), and red (undesired class). The factual, $x$, moves over the sequence, as do the respective target counterfactual points (stars). Between segments of the true trajectory (e.g. $x_1$, $x_2$) TraCE measures alignment in angle, $R_1$, and the ``best move'' given the angle, $R_2$, with respect to counterfactual target points (stars in the left panel).  In this example the TraCE score for moving from $x_0$ to $x_1$ is negative (-0.1855) because it aligns more with the negative counterfactual (red class), whereas the trajectory from $x_1$ to $x_2$ is away from the negative counterfactual and towards the positive counterfactual (blue class) hence the positive score (0.4056).}
    \label{fig:toy}
\end{figure*}

Many counterfactual explainers have been developed and most commonly they are applied to single-step decision making processes involving one data point per individual~\cite{guidotti2022counterfactual}. Relatively limited research has been conducted into more complex counterfactual techniques and applications for sequential and time series applications. Such research has mostly focused on counterfactuals in the context of multivariate time series explainability~\cite{ates2021counterfactual, hollig2022tsevo}, recourse as a sequence of actions~\cite{tsirtsis2021counterfactual}, and suggested alterations to particular regions of an individual time series~\cite{delaney2021instance}.

We hypothesise that counterfactual explanations could provide insights beyond their current role in the development of explainable systems by utilising them as benchmarks to evaluate trajectories or sequences of decisions. To this end, we introduce TraCE (Trajectory Counterfactual Explanation) scores, which consider the sequence of steps in a task and compare each step to counterfactual examples, including both desirable and undesirable targets. In the example of the intensive care unit patient, at each point in the patient's stay, TraCE's objective is to evaluate the true trajectory against a potential path towards survival (desirable counterfactual), and mortality (undesirable counterfactual). TraCE scores aim to provide an easily understandable sequential assessment of trajectory, enabling progress tracking in a specified task for laypeople and domain experts alike.

\section{Preliminaries}

Counterfactual explanations are often used to assess what actions are required to push a query point (the factual) over the decision boundary of a model in order to produce a different outcome (the counterfactual)~\cite{wachter2018a}. Adversarial examples stand in stark contrast to counterfactual explanations, as they explicitly seek to misclassify the factual by deceitfully perturbing its features~\cite{goodfellow2014explaining}. 

In essence, counterfactual explanations encapsulate the thought experiment:
\begin{center}
    \textit{$Y$ was my outcome, but if I had done $Z$ \\ then $Y^\prime$ would have occurred instead.}
\end{center}
Therefore, given a decision maker $f$, a set of possible outcomes $\{y, y\prime\}$ and a query point $x$, a counterfactual looks like:
\begin{equation*}
    f(x) = y, \quad f(x + z) = y^\prime
\end{equation*}
where $z$ is the change on $x$ in order to achieve $y^\prime$. In the hospital example, where $x$ is the patient, $y$ is their predicted outcome (for example mortality), $y\prime$ is the counterfactual representing an alternative outcome (for example successful discharge), and $z$ is the set of feature changes required to lead to this alternative outcome. We can constrain $z$ to fulfill certain criteria, such as minimising complexity (sparse $z$) or length (small $\lVert z \rVert$)~\cite{VIRGOLIN2023103840}, maximising feasibility (follow probability distributions)~\cite{poyiadzi2020face} or agency (follow multiple possibilities)~\cite{sokol2023navigating}.

\paragraph{Notation} We define scalar values as Greek letters e.g., $\alpha$, and an input space $\mathcal{X}$ without loss of generality. That is to say, $\mathcal{X}$ can take the form of a set of one-dimensional features (vector), image space, a compressed/latent space, etc. We require that $\mathcal{X}$ is a real vector space with a well-defined inner product $ \langle \cdot \; , \; \cdot \rangle : \mathcal{X} \times \mathcal{X}  \mapsto \mathbbm{R}$ which follows the usual properties. The inner-product induced norm is defined as $\lVert v \rVert = \sqrt{\langle v \; , \; v \rangle}$. We could also use a sensible distance function $d:\mathcal{X}\times\mathcal{X}\mapsto \mathbbm{R}_+$ which must follow the usual axioms of a distance function.

We take $x_t\in\mathcal{X}$ as a singular instance taken at time $t$ from the input space, with $x_{t}^\prime$ to be the target point associated with $x_t$. $x^\prime$ can be defined using any arbitrary process, e.g. a counterfactual generated with a model or a goal set by a domain expert. We then define the true change, $v_t$, and the desired change, $v_t^\prime$:
\begin{equation}
    \begin{aligned}
        v_t = x_{t+1} - x_{t}, \quad
        v_{t}^\prime = x_{t}^\prime - x_t
    \end{aligned}
\end{equation}

\begin{theorem}
    Given $a,b,c\in\mathbbm{R}^n$, the closest point $d$ to $a$ in the vector direction $c-b$ is:
    \begin{equation}
    \label{eq:close}
         d = b + \frac{h}{\lVert h \rVert}\cdot\lVert g \rVert \cdot \theta
    \end{equation}
    where $h=c-b$, $g=a-b$ and $\theta = \frac{\langle h \; , \; g \rangle}{\lVert h \rVert \lVert g \rVert}$.
\end{theorem}
Proof in Appendix~\ref{app:proof}.

\section{TraCE}

Trajectory Counterfactual Explanation (TraCE) scores $S:\mathcal{X}\times\mathcal{X}\mapsto[-1,1]$ condense the complex task of tracking progress towards successive counterfactual targets through time into a single number between $-1$ and $1$. This single number requires no expertise or domain knowledge to interpret. Simply put:
\begin{itemize}
    \item $S<0$ implies that $x_{t+1}$ is further from $x_t^\prime$ than $x_{t}$, with $S\to-1 \implies \lVert x_t^\prime - x_{t+1} \rVert \gg \lVert x_t^\prime - x_{t} \rVert$. For the hospital patient example, when applied to a desirable counterfactual, $S<0$ implies that the patient is moving further from the desired region (discharge) and is deteriorating;
    \item $S>0$ implies that $x_{t+1}$ is closer to $x_t^\prime$ than $x_{t}$, with $S\to1 \implies \lVert x_t^\prime - x_{t+1} \rVert \ll \lVert x_t^\prime - x_{t} \rVert$, suggesting that the patient is improving and getting closer to successful discharge; and
    \item $S=0$ implies no movement towards or away from a target, so $\lVert x_t^\prime - x_{t+1} \rVert = \lVert x_t^\prime - x_{t} \rVert$, suggesting that the patient is neither getting better or worse relative to the counterfactual target(s).
\end{itemize}

In order to do this, we track two metrics: (1) the angle between the real change and the desired change $R_1(x_t, x^\prime_t)$; and (2) the distance travelled relative to the angle $R_2(x_t, x^\prime_t)$.

The angle between the true trajectory and desired trajectory can simply be measured using the normalised dot product:
\begin{equation}
    \label{eq:norm_dot}
    R_1(x_t, x^\prime_t) = \frac{\langle v_t \; , \; v^\prime_t \rangle}{\lVert v_t \rVert \lVert v^\prime_t \rVert}
\end{equation}
 From Theorem 1, given the angle score $R_1(x_t, x^\prime_t) = \theta_t$, if $\theta_t > 0$ then the closest point $\hat{x}_t$ to $x^\prime_t$ is:
\begin{equation*}
    \hat{x}_t = x_t + \frac{v_t}{\lVert v_t \rVert}\lVert v^\prime_t \rVert \theta_t
\end{equation*}
whereas if $\theta_t \leq 0$ the distance from $x^\prime_t$ is increasing, and so $\hat{x}_t = x_t$. Thus:
\begin{equation}
    \label{eq:landing}
    R_2(x_t, x^\prime_t) = \Big\lvert\frac{\langle\hat{v}_t \; , \;v^*_{t} \rangle}{ \lVert \hat{v}_t \rVert \lVert v^*_{t}\rVert} \Big\rvert
\end{equation}
where:
\begin{equation*}
    \hat{v}_t = x^\prime_t - \hat{x}_t, \quad v^*_{t} = x^\prime_{t} - x_{t+1}
\end{equation*}
Thus $R_2=1$ when $x_{t+1}=\hat{x}_t$. We then combine Equations~\ref{eq:norm_dot} and~\ref{eq:landing} into a single score:
\begin{equation}
    S(x_t, x^\prime_t) = \lambda R_1(x_t, x^\prime_t) + (1 - \lambda) R_2(x_t, x^\prime_t)
\end{equation}
where $\lambda\in[0,1]$ is a weight which can be either a scalar value or a function.

TraCE can consider progress towards a single class (as presented in Section~\ref{SSP}), or multiple classes encompassing both desirable and undesirable counterfactuals (Section~\ref{ICU}). Figure~\ref{fig:toy} encapsulates the latter scenario, where we assess progress towards two classes, one desirable and one undesirable, via an average between measured progress towards each outcome as the factual changes. Here we can see that if the distance between sequential factual instances is small, and/or if two counterfactual points from different classes are in close proximity (relative to their distance from the factual), it can be difficult to assess how any change to the factual may contribute to the final outcome. TraCE addresses this. $\lambda>\frac{1}{2}$ implies we care more about the trajectory angle than the distance travelled. When $\lambda \neq 1$, $S=1$ implies $x_{t+1}=x^\prime_t$, and so the goal has been achieved. Code is available to implement TraCE \footnote{\url{https://github.com/jeffnclark/TraCE}}.

\section{Case Studies}

Here we demonstrate the use of TraCE scores in two real-world case studies.

\subsection{Intensive care unit outcomes}
\label{ICU}

\begin{figure*}[t]
     \centering
          \begin{subfigure}[t]{0.48\linewidth}
         \centering
             \includegraphics[width=\linewidth]{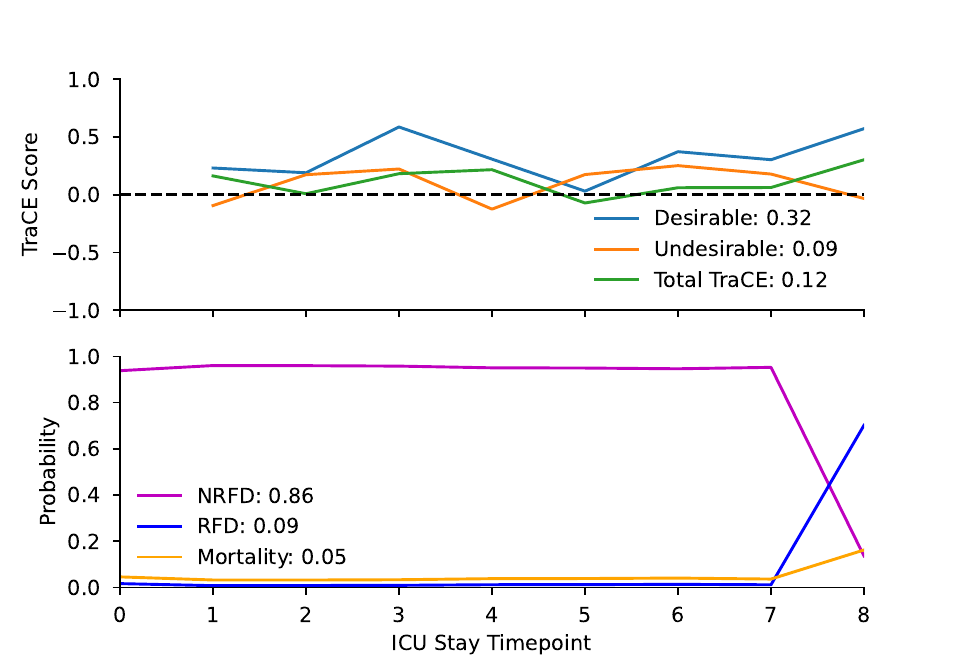}
          \caption{Successfully discharged patient trajectory}
          \label{fig:ICU_pos}
     \end{subfigure}
     \hfill
     \begin{subfigure}[t]{0.48\linewidth}
         \centering
             \includegraphics[width=\linewidth]{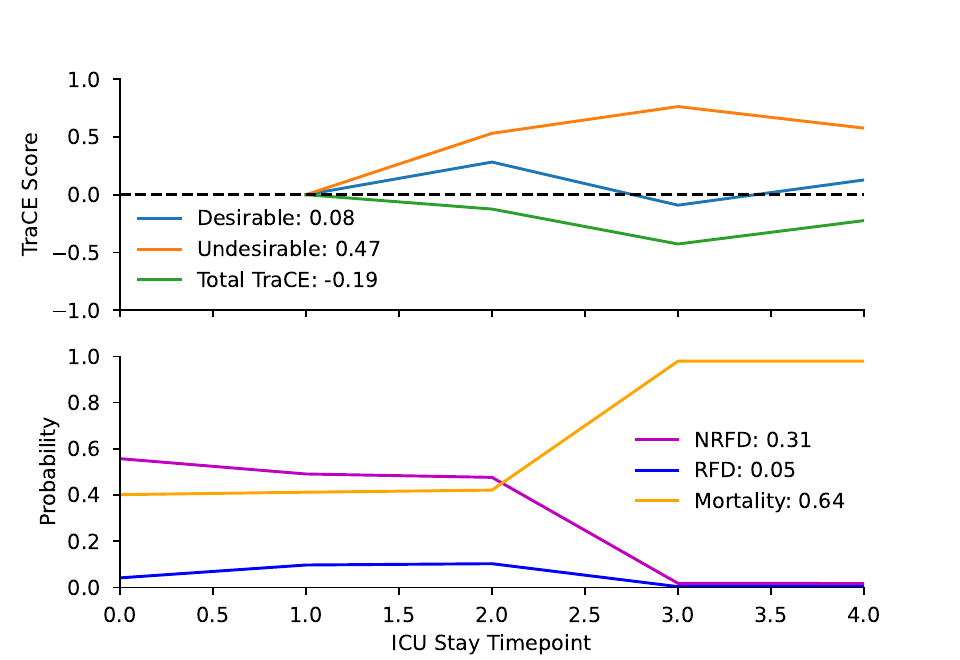}

         \caption{
    In-hospital mortality patient trajectory} \label{fig:ICU_neg}
     \end{subfigure}

     \caption{Contrasting example patient journeys. For each, top: instantaneous TraCE scores, higher indicates more alignment with the specified counterfactuals. `Desirable' refers to alignment with successful discharge counterfactuals, `Undesirable' refers to mortality counterfactuals. TraCE is computed on the current and preceding time point, hence time point 0 is not presented. Bottom: Classifier probabilities via the prediction model. Values in the legends are averages across the whole trajectory. NRFD = Not ready for discharge, RFD = ready for discharge (desirable outcome), mortality (undesirable outcome).}
     \label{fig:ICU_main}

\end{figure*}

Clinical care involves a huge number of dynamic variables which must be considered when making decisions. Clinical scores, such as APACHE and NEWS, are widely used to provide a snapshot of a patient's current status relative to established benchmarks~\cite{gerry2020early}. However, these scores fail to capture dynamics and lack personalization to a patient's scenario. TraCE is able to overcome these shortcomings in existing clinical scores by better capturing the dynamic progress of an individual patient. Here we demonstrate the application of TraCE to intensive care unit (ICU) patients, relative to counterfactuals for successful discharge and in-hospital mortality.

\subsubsection{Methods}
Time series intensive care unit data were extracted from the MIMIC IV 2.0 data set~\cite{mimicIV_20}. Seventeen features, including vital signs such as heart rate and respiratory rate, were identified for TraCE, following existing research~\cite{mcwilliams2019towards}. Outcome labels were generated using known outcomes, for discharge to home or mortality. Patients discharged to locations other than home were removed, leaving a total of 327270 time points across 30860 hospital stays (26089 patients) for analysis. All time points prior to the final time point were labelled as not ready for discharge. Missing proceeding data in the time series were completed using forward fill and, for missing initial values, backward fill. Numerical features empty across each patient's whole stay were filled with the class average, while absent categorical features were filled with the class mode. All features were normalised.

Using scikit-learn, a multi-layer perceptron  classifier, with two hidden layers of 10 neurons each, was trained for a maximum of 10 epochs on individual patient time steps to predict three classes: not ready for discharge, ready for discharge, mortality. All other hyperparameters were as default. Classes were balanced by undersampling and an 80:20 train:test split was utilised.

TraCE analysis was carried out as follows for 1000 hospital stays in the test set, 500 known to be successfully discharged to home, 500 unsuccessfully discharged patients (in-hospital mortality). KDTrees for each outcome class were generated from the corpus of known outcomes within the training set. For each time step in a patient's hospital stay, counterfactuals (n = 3) were sampled from each KDTree, resulting in ready for discharge (desired) counterfactuals and in-hospital mortality (undesired) counterfactuals. TraCE was implemented ($\lambda = 0.9$) against each of these counterfactuals and compared with class probabilities calculated by the classifier. Static features which did not differ between the factual and counterfactual were omitted from TraCE analysis, as were time steps where no features changed. Welch's t-test was performed to test if average TraCE scores differed between the two outcome groups.

\subsubsection{Results and Discussion}

The multilayer perceptron classifier achieved test set accuracy of 0.95. The average TraCE score for 500 patients known to be successfully discharged to home was 0.0821 (SD 0.1373). For 500 unsuccessfully discharged patients (in-hospital mortality), their average TraCE score was -0.0302 (SD 0.0675). The difference in average TraCE score was statistically significant ($p < .00001$). Since a patient is typically not ready for discharge (NRFD) for most of the stay, an average near $0$ is expected. More intelligent weighting of variables, coupled with expertise provided by clinicians, is likely to further increase the TraCE score gap between patients with positive and negative outcomes.

Instantaneous TraCE score values between successive time points are expected to be more useful at potential deployment than average scores, and plots for which are shown in Figure \ref{fig:ICU_main} for two patients with different outcomes.

TraCE scores plotted for a patient successfully discharged to home show signs of positive progress towards discharge early in the stay, as indicated by the high alignment with desirable counterfactuals (Figure \ref{fig:ICU_pos}, top). The MLP classifier does not capture this progress, with stable probabilities for all three classes until the final timepoint (patient discharge), and in fact higher likelihood of mortality than readiness for discharge for the majority of the ICU stay (Figure \ref{fig:ICU_pos}, bottom). An additional example trajectory of successful discharge can be found in Appendix \ref{fig:ICU_pos_supp}. For cases such as these, real-time observation of TraCE scores could provide early insights into patient improvement. 

We also present a negative outcome ICU stay which resulted in in-hospital mortality (Figure \ref{fig:ICU_neg}). For the first half of the stay the classifier most likely predicts not ready for discharge (NRFD) closely followed by mortality. The high mortality probability is reflected in the instantaneous TraCE score which aligns more with the undesirable (mortality) counterfactual than the desirable (ready for discharge), and negative trend in total TraCE score. Patient deterioration is indicated by the TraCE scores at timepoint 2 (increasing undesirable TraCE component) whereas the classifier does not increase the risk of mortality until timepoint 3. Plots for an additional negative outcome patient trajectory is presented in Appendix \ref{fig:ICU_neg_supp}. In instances of patient decline, early intervention is critical and TraCE may provide additional insights to compliment existing tools.

TraCE enables determination of the optimal vector for any single time point which would maximise the TraCE score by considering not just positive alignment with the desired outcome but also negative alignment with the undesired outcome. In a clinical setting, this insight could be applied prospectively, by suggesting optimal actions for a current patient in ICU. Likewise, clinicians are able to specify desirable and undesirable counterfactual targets which could be personalised for a given patient. For example, if it may not be reasonably expected that a patient will make a full recovery, the desirable counterfactual could be adjusted to match expectations such as discharge to a nursing facility.

With refinement, the presentation of TraCE scores in a clinical dashboard could provide clinicians with a digestible real-time summary of patient progress. Future work in developing TraCE for this application, such as weighting TraCE to certain events, analysing the gradient and stability of TraCE scores during the ICU stay and considering counterfactual path feasibility, may yield an improved understanding of a patient's health trajectory to inform and improve quality of care.

\subsection{Monitoring sustainable global development}
\label{SSP}

To address the ongoing climate emergency, it is critical to reconcile global socioeconomic development with environmental sustainability. However, it is difficult to holistically evaluate a region's overall development trajectory, due to multifaceted social, economic, and environmental considerations. In 2017, five development narratives were published in the form of Shared Socioeconomic Pathways (SSPs): (1) Sustainability, (2) Middle of the Road, (3) Regional Rivalry, (4) Inequality, (5) Fossil-fueled Development \cite{Riahi2017_oecd_ssps, ONEILL2017169_SSPs}. These characterise changing socioeconomic factors for the next century, and the associated changes in emissions of greenhouse gases and air pollutants. In this application, TraCE quantifies the overall development sustainability of different countries, relative to each of these established SSP scenarios, with a view to monitoring alignment with the development trajectories to date.

\subsubsection{Methods}
Global time series data for socioeconomic and environmental features was extracted for the years 2015-2022. For the environmental features (surface temperature, precipitation, methane concentration), ERA5 reanalysis data \cite{era5data_historical_climate} and satellite data \cite{methane_historical_atmospheric} were used to represent the factual historical features, and counterfactuals were represented by CMIP6 projections for the baseline scenario of each SSP \cite{cmip6data_ssp_climate, methane_ssp1, methane_ssp2, methane_ssp3, methane_ssp4, methane_ssp5}. Factual and counterfactual representations for the socioeconomic features (population, GDP) were similarly obtained from OECD historical datasets \cite{oecd_historical_population, oecd_historical_gdp} and SSP projections \cite{Riahi2017_oecd_ssps, Samir2017_oecd_ssps_population, Dellink2017_oecd_ssps_gdp} respectively. To address differences in spatiotemporal resolutions, spatial coverage, and missing data points in the datasets, the chosen feature data was aggregated to monthly mean values and normalised at the country level, resulting in features for 34 countries. For each SSP, TraCE scores were calculated ($\lambda = 0.9$) between the actual feature data and the matching monthly SSP projection data as the target point. No undesirable counterfactual point was assigned. TraCE scores for each SSP were then compared, to quantify the alignment of a given country's development trajectory with the different SSPs.

\begin{figure}[t]
    \centering
    \includegraphics[width=0.48\textwidth]{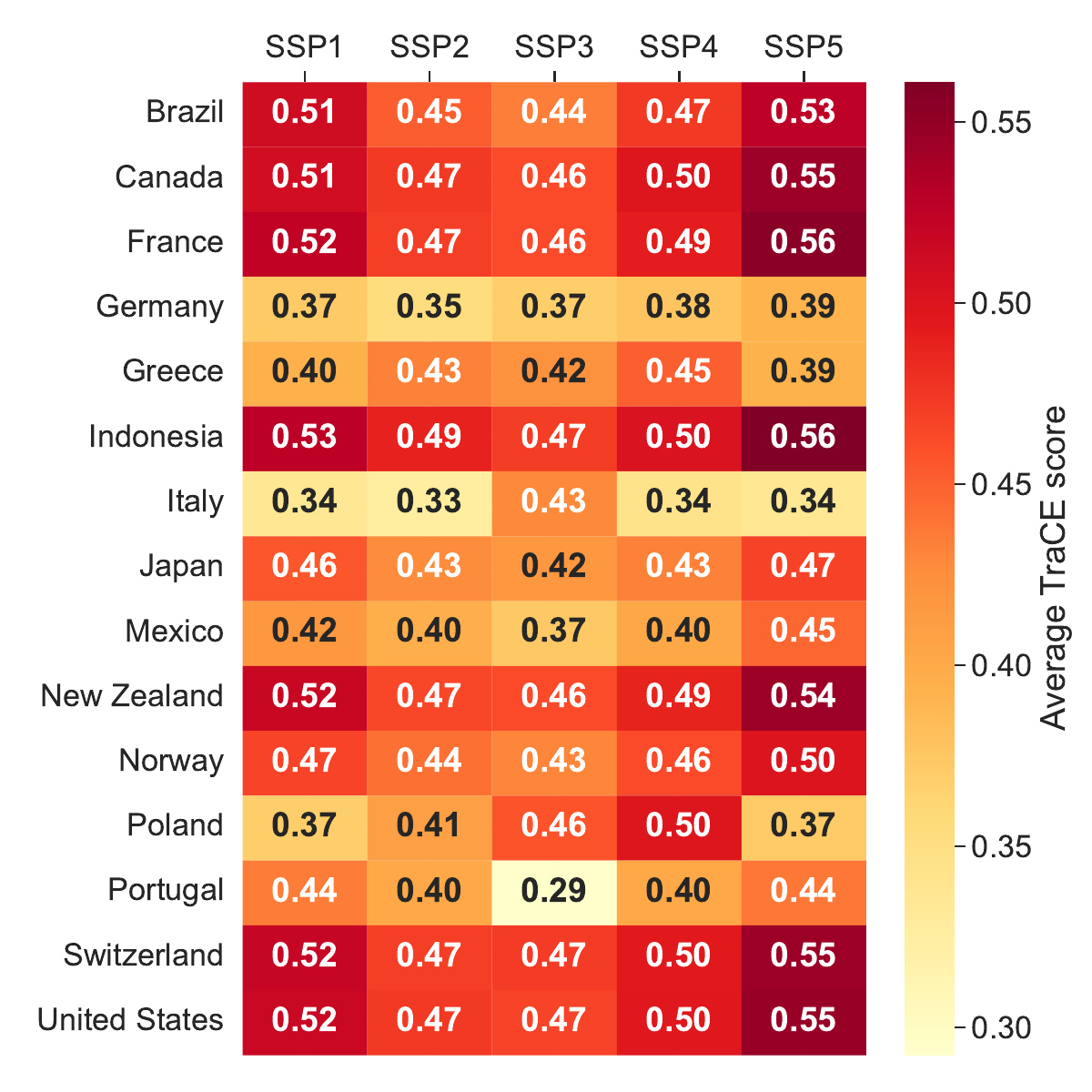}
    \caption{Average TraCE scores for each SSP for 15 countries, across the period 2015-2022. Higher TraCE scores indicate closer alignment with a Shared Socioeconomic Pathway (SSP).}
    \label{fig:ssp_trace_heatmap}
\end{figure}

\subsubsection{Results and Discussion}
Analysis of the average TraCE scores for 15 different countries found that most countries in the study fit a common pattern.  An overview of the countries' alignments with SSP projections is shown in the heatmap (Figure \ref{fig:ssp_trace_heatmap}) for the study period 2015-2022. Comparisons can be made between SSPs for a single country, and across different countries. A common pattern emerges across most countries, with SSP5 (Fossil-fueled Development) ranking highest, followed by SSP1 (Sustainability), closely tracked by SSP4 (Inequality), and finally, SSP2 (Middle of the Road) and SSP3 (Regional Rivalry). Some notable results stand out: several countries, including Germany, Greece, Italy, Mexico, and Portugal, exhibit lower TraCE scores across all SSPs. This indicates that their observed data features are less similar to their corresponding SSP projections, when compared to most other countries in the study. Additionally, some countries deviate from the majority SSP ranking pattern. For example, Greece aligns most closely with SSP4, followed by SSP2 and SSP3, with SSP1 and SSP5 ranking the lowest. Italy aligns most with SSP3, showing strong divergence from the remaining SSPs, which have similar TraCE scores. Poland closely aligns with SSP4, followed by SSP3, with TraCE scores diverging significantly from the other SSPs.

Importantly, this work does not provide evidence for attributing specific actions or responsibility to particular countries. This is because the observed data features for a given country can be influenced by the actions of other countries. Instead, TraCE scores can serve as a monitoring metric, or an output metric in simulation experiments, because they quantify the alignment of observed data features with SSP projections.

\begin{figure}[t]
    \centering
    \includegraphics[width=0.48\textwidth]{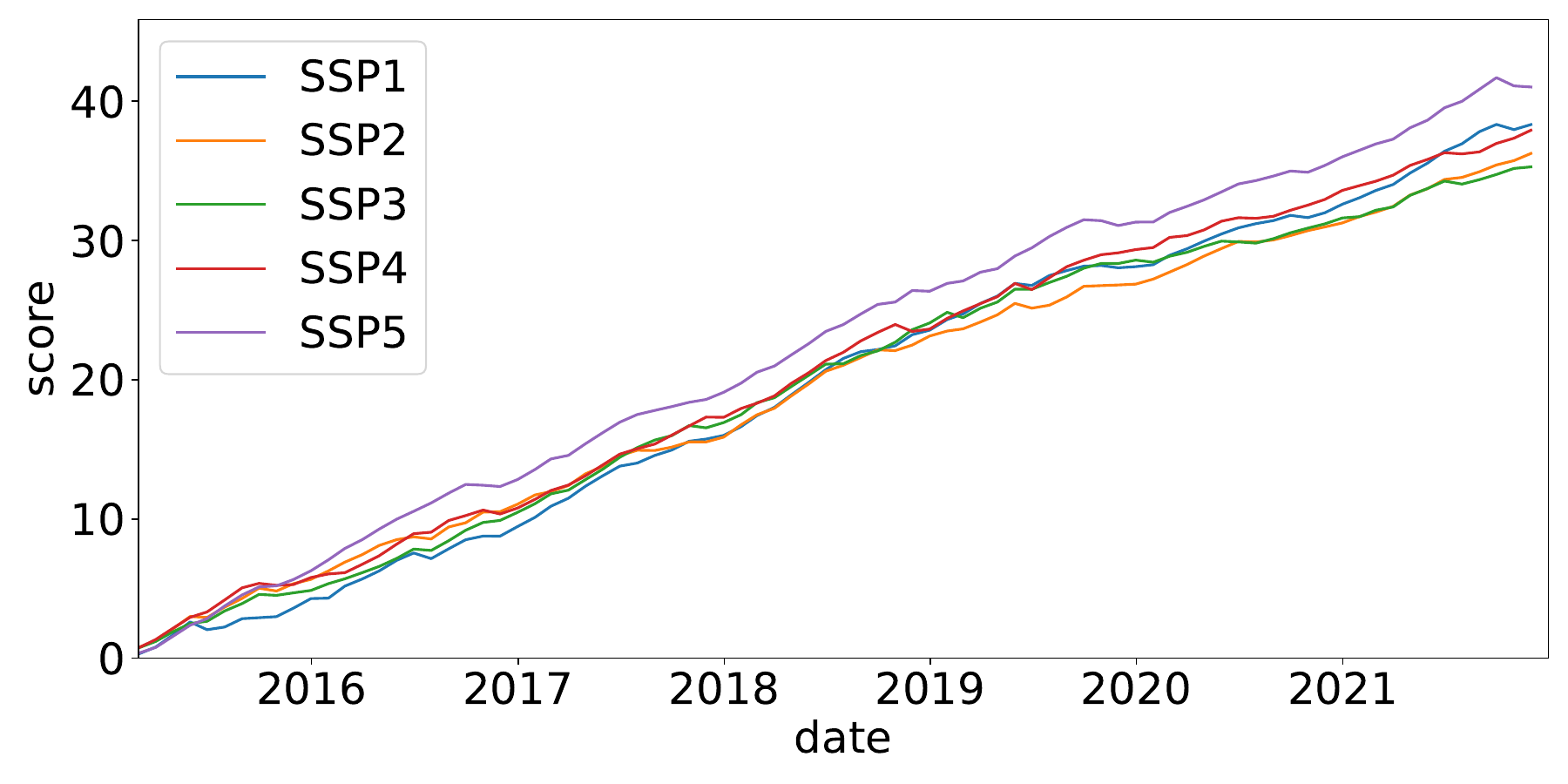}
    \caption{Cumulative monthly SSP TraCE scores for Norway, 2015-2022.}
    \label{fig:ssp_trace_timeseries_cumulative}
\end{figure}

Figure \ref{fig:ssp_trace_timeseries_cumulative} shows the cumulative TraCE score time series (2015-2022) for Norway, which was identified as a representative country. The TraCE score trajectories are consistently positive across all SSPs, in agreement with the expectation that they were developed as realistic scenarios in alignment with the factual historical data. Of note is the visible flattening around the year 2020, which coincides with the onset of the COVID-19 pandemic. This flattening likely occurs because the SSP projections did not anticipate the pandemic, so the observed data features deviate from these trajectories, resulting in low or negative instantaneous TraCE scores. Overall, Figure \ref{fig:ssp_trace_timeseries_cumulative} indicates that SSP5 consistently ranks the highest from 2016 onwards, while other SSPs score more closely together. However, starting in mid-2021, the SSP4 and SSP1 TraCE scores begin to diverge above those of SSP2 and SSP3. With refinement, future work could correlate temporal TraCE scores with societal events and political decisions, such as legislation. Additional plots presenting the findings for Poland, as a contrasting example, are available in Appendix \ref{app:ssp_outliers}, including a heatmap of feature importance to provide preliminary explainability of TraCE.

It must be emphasised that this study serves as a proof of concept, and requires input from experts across multiple domains to ensure safe and trustworthy implementation. This includes the selection of data features for monitoring, and their weighting, which has been equally distributed in this demonstration. Different weighting schemes will yield distinct results and should be developed in accordance with the priorities and specific questions of the user. Additionally, the data used and results obtained are contingent on the model source for SSP projections. 

The utility of TraCE scores in this application lies in the capability to reconcile complex and occasionally conflicting variables into a single value. This allows experts and non-experts to quickly assess alignment with the established SSP scenarios, via an explainable method based on direction and distance in the data feature space. Visually assessing such alignment from the raw data itself can be challenging, particularly as the number of included features increases. The TraCE method is therefore useful for communication and understanding between stakeholder groups, and with refinement could aid monitoring of region sustainability against established development pathways.

\section{Conclusion}
TraCE provides a model-agnostic modular framework from which to assess progress over time towards an assigned goal. As demonstrated, the modularity of TraCE enables application-specific adaptation. Counterfactual target points can be defined as most appropriate, such as: model-generated counterfactuals, corpus of examples, expert-selected landmarks, or industry benchmarks.

The presented case studies involve at most 17 features. TraCE's utility is expected to become even more evident with higher complexity scenarios which likely involve larger neural networks. In this paper we present TraCE scores in several forms: instantaneous (ICU study, Section \ref{ICU}); average and cumulative (SSP study, Section \ref{SSP}). More sophisticated methods to harness the temporal dimension could be considered after calculating TraCE scores such as quantifying instability, gradients through successive time steps, or time-dependent score weighting. The implementation of TraCE for the presented applications are for illustrative purposes, deployment and interpretation of TraCE should be guided by domain experts. Further work is required for robust implementation, including feature selection and tuning of $\lambda$.

By distilling high dimensional dynamic sequential tasks into a single value, TraCE scores enable experts and laypeople alike to quantify and better understand progress in sequential tasks.

\section{Acknowledgements}
We thank Thea Barnes for SQL scripts for MIMIC IV data extraction.
JNC, MWLW and RSR are funded by the UKRI Turing AI Fellowship [grant number EP/V024817/1]. EAS is funded by the ARC Centre of Excellence for Automated Decision-Making and Society (project number CE200100005), funded by the Australian Government through the Australian Research Council. EFM is funded by a Google PhD Fellowship. Part of this work was done within the University of Bristol’s Machine Learning and Computer Vision (MaVi) Summer Research Program 2023.

\printbibliography

\newpage

\appendix
\section{Appendix}

\subsection{Proof of Theorem 1}
\label{app:proof}
\paragraph{Claim}
Given $a,b,c\in\mathbbm{R}^n$, the closest point $d$ to $a$ in the vector direction $c-b$ is:
    \begin{equation*}
         d = b + \frac{h}{\lVert h \rVert}\cdot\lVert g \rVert \cdot \theta
    \end{equation*}
    where $h=c-b$, $g=a-b$ and $\theta = \frac{\langle h \; , \; g \rangle}{\lVert h \rVert \lVert g \rVert}$.

\begin{proof}
\begin{figure}[ht]
    \centering
    \includegraphics[width=0.25\textwidth]{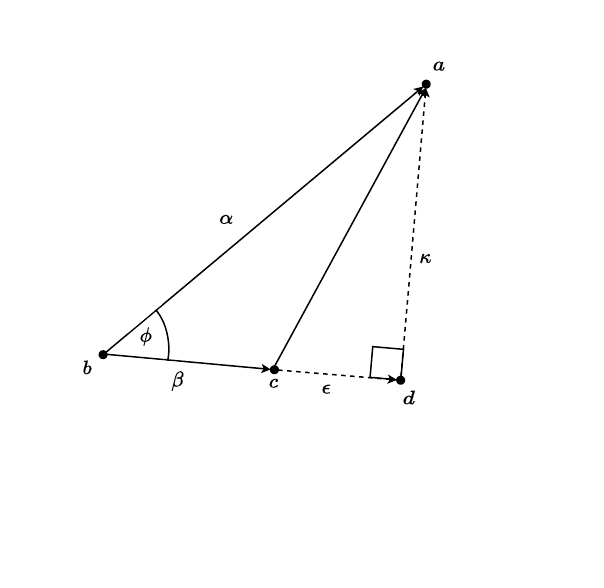}
    \caption{Geometric image of the proof for Theorem 1.}
    \label{fig:tri}
\end{figure}
    In $n$-dimensional space, the points $a, b, c\in\mathbbm{R}^n$ create a triangle. Define $\alpha = \lVert g \rVert = \lVert a - b \rVert$ and $\beta = \lVert h \rVert = \lVert c - b \rVert$. Since the closest point along a line to another point must form a perpendicular vector, for $d$ to be the closest point along the vector direction $c - b$, $a, b, d$ must form a right angled triangle, shown in Figure~\ref{fig:tri}. Thus, define $\epsilon = \lVert d - c \rVert$, $\kappa = \lVert a - d \rVert$, from Pythagoras Theorem:
    \begin{equation*}
    \begin{aligned}
        (\beta + \epsilon)^2 + \kappa^2 &= \alpha^2 \\
        \implies (\beta + \epsilon) &= \sqrt{\alpha^2 - \kappa^2}
    \end{aligned}
    \end{equation*}
    From trigonometric identities, $\kappa = \alpha\sin(\phi)$ and
    \begin{equation*}
        \phi = \arccos\bigg(\frac{\langle h \; , \; g \rangle}{\lVert h \rVert \lVert g \rVert}\bigg)
    \end{equation*}
    thus:
    \begin{equation*}
        \kappa = \alpha\sqrt{1 - \frac{\langle h \; , \; g \rangle}{\lVert h \rVert \lVert g \rVert}^2}
    \end{equation*}
    giving:
    \begin{equation*}
        (\beta + \epsilon) = \sqrt{\alpha^2 - \bigg(\alpha\sqrt{1 - \frac{\langle h \; , \; g \rangle}{\lVert h \rVert \lVert g \rVert}^2}\bigg)^2}
    \end{equation*}
    Since the normalised dot product is strictly $[-1, 1]$:
    \begin{equation*}
        0 \leq \sqrt{1 - \frac{\langle h \; , \; g \rangle}{\lVert h \rVert \lVert g \rVert}^2} \leq 1
    \end{equation*}
    therefore:
    \begin{equation*}
        \alpha \geq \alpha\sqrt{1 - \frac{\langle h \; , \; g \rangle}{\lVert h \rVert \lVert g \rVert}^2}
    \end{equation*}
    and so $\beta + \epsilon\in\mathbbm{R}_+$, giving:
    \begin{equation*}
    \begin{aligned}
            (b+\epsilon) &= \sqrt{\alpha^2 - \bigg(\alpha\sqrt{1 - \frac{\langle h \; , \; g \rangle}{\lVert h \rVert \lVert g \rVert}^2}\bigg)^2} \\
            &= \sqrt{\alpha^2 - \alpha^2\bigg(1 - \frac{\langle h \; , \; g \rangle}{\lVert h \rVert \lVert g \rVert}^2\bigg)} \\
            &= \sqrt{\alpha^2 \bigg(\frac{\langle h \; , \; g \rangle}{\lVert h \rVert \lVert g \rVert} \bigg)^2} \\
            &= \alpha \frac{\langle h \; , \; g \rangle}{\lVert h \rVert \lVert g \rVert}
    \end{aligned}
    \end{equation*}
    $\beta + \epsilon$ describes the distance we must travel along the vector direction $c-b$ to get from $b$ to $d$. Therefore:
    \begin{equation}
        d = b + \frac{h}{\lVert h \rVert}(\beta + \epsilon)
    \end{equation}
    which gives Equation~\ref{eq:close} when substitution is complete.
\end{proof}

\subsection{Intensive care unit outcomes}
\label{app:icu}

\begin{figure*}[t]
    \label{fig:ICU_supp}
     \centering
          \begin{subfigure}[t]{0.48\linewidth}
         \centering
             \includegraphics[width=\linewidth]{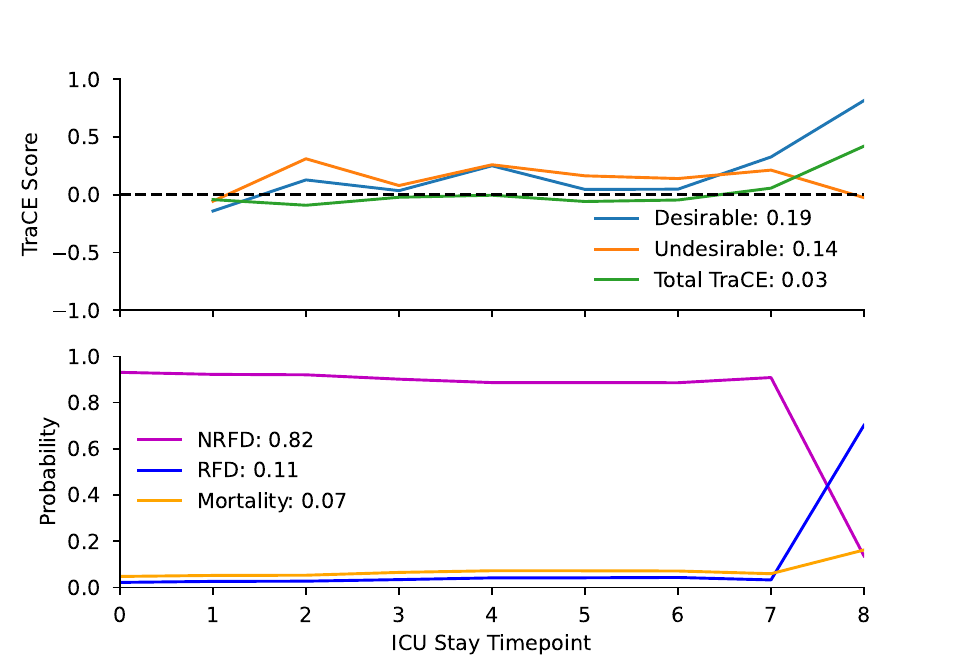}
          \caption{Successfully discharged patient trajectory}
          \label{fig:ICU_pos_supp}
     \end{subfigure}
     \hfill
     \begin{subfigure}[t]{0.48\linewidth}
         \centering
             \includegraphics[width=\linewidth]{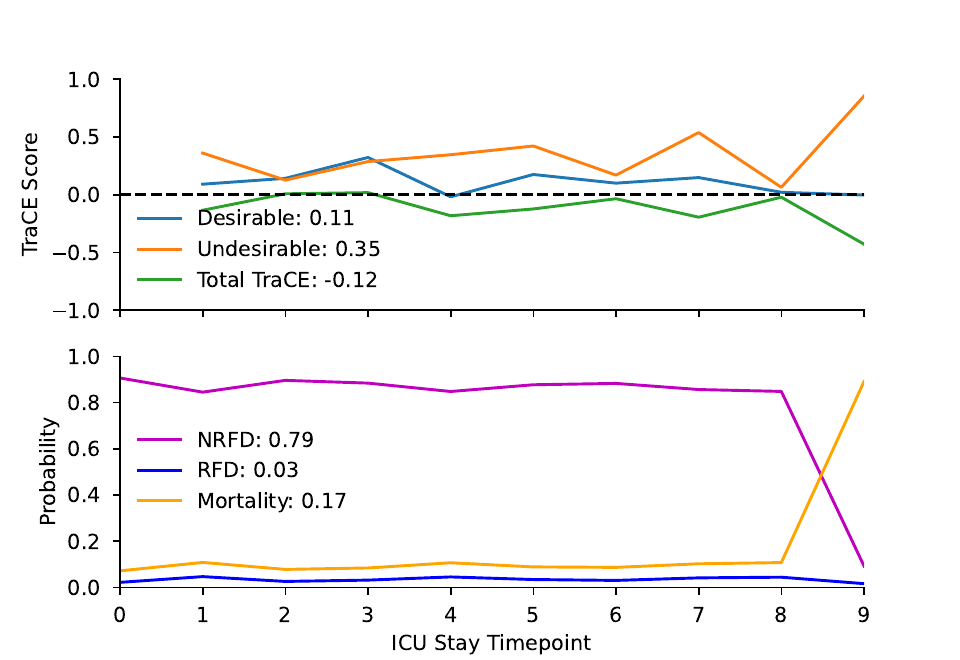}

         \caption{
    In-hospital mortality patient trajectory} \label{fig:ICU_neg_supp}
     \end{subfigure}

     \caption{Contrasting example patient journeys. For each, top: Instantaneous TraCE scores, higher indicates more alignment with a counterfactual. `Desirable' refers to alignment with successful discharge counterfactuals, `Undesirable' refers to mortality counterfactuals. TraCE is computed on the current and preceding time point, hence time point 0 is not presented. Bottom: Classifier probabilities via prediction model. Values in the legends are averages across the whole trajectory. NRFD = Not ready for discharge, RFD = ready for discharge (desirable outcome), mortality (undesirable outcome).}
\end{figure*}

Figure \ref{fig:ICU_pos_supp} shows TraCE applied to another ICU patient who was successfully discharged to home. For the first two-thirds of the stay, the patient's predicted probability of mortality was higher than for successful discharge (RFD), which is reflected by the stronger alignment with the undesirable counterfactuals (mortality) in this portion of the stay. However, the patient does recover and goes on to be successfully discharged. The TraCE score begins to increase (timepoint 7) prior to the patient's improved health being reflected in the classifier probabilities (timepoint 8).

An unsuccessfully discharged ICU patient is shown in Figure \ref{fig:ICU_neg_supp}. In this case from the TraCE score it is evident throughout the stay that the patient is deteriorating, given the consistently higher alignment with the undesirable (mortality) counterfactuals than the desirable (discharged to home) counterfactuals. This demonstrated the patient's increasing proximity to the undesirable outcome, mortality. The increasing risk of mortality is not reflected by the classifier (Figure \ref{fig:ICU_neg_supp}, bottom), which is not apparent until the patient's final timepoint. Until this point the probability plot appeared very similar to the previously described patient (Figure \ref{fig:ICU_pos_supp}). This suggests that with refinement TraCE could provide utility, as part of a clinician's toolkit, to support decisions and ultimately improve patient care.

\subsection{Monitoring sustainable global
development}
\label{app:ssp_outliers}
\begin{figure}[ht]
    \centering
    \includegraphics[width=0.49\textwidth]{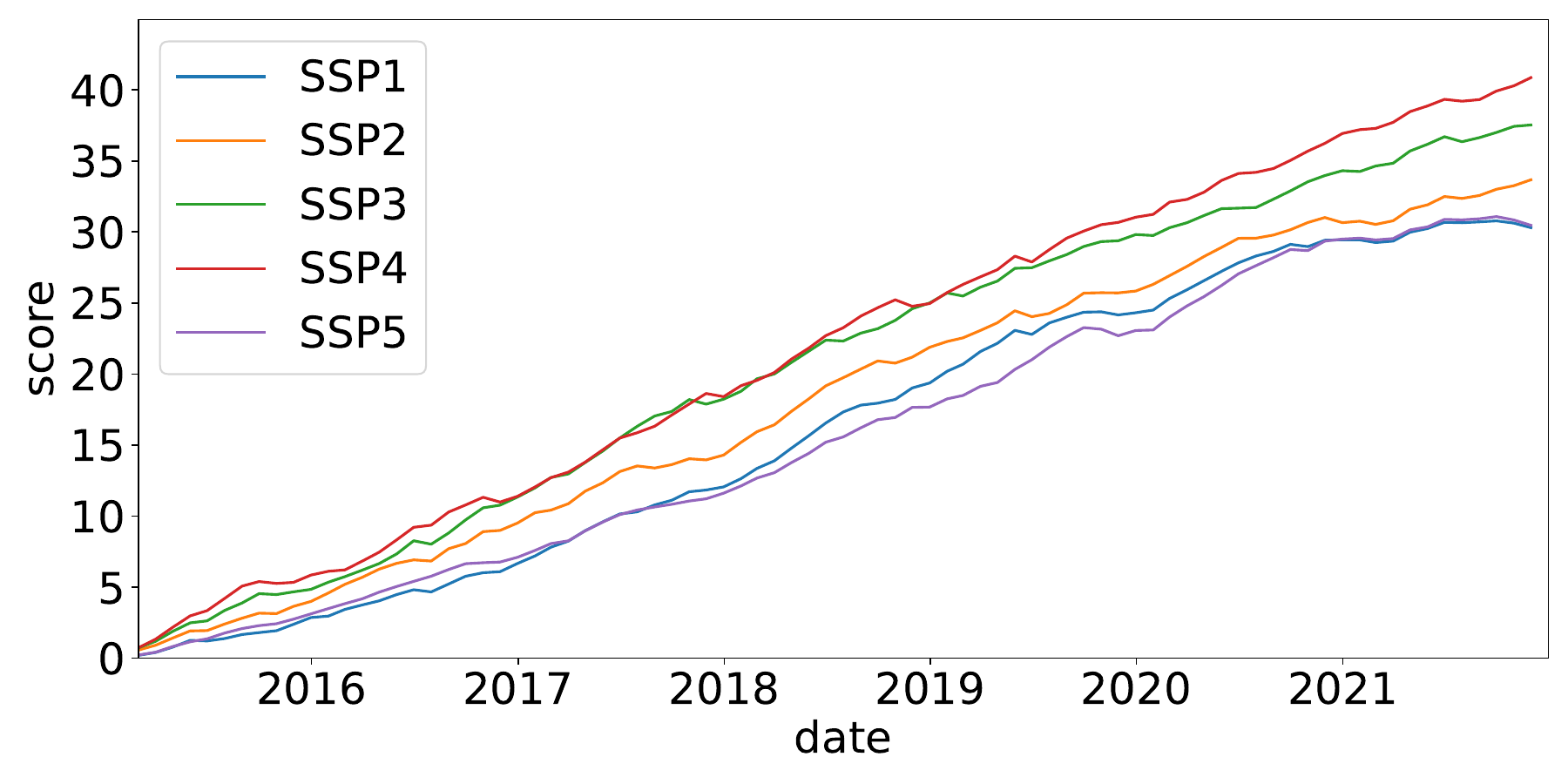}
    \caption{Cumulative monthly TraCE scores for Poland, 2015-2022. Higher TraCE score indicates closer SSP alignment.}
    \label{fig:appendix_trace_timeseries}
\end{figure}

\begin{figure}[ht]
    \centering
    \includegraphics[width=0.49\textwidth]{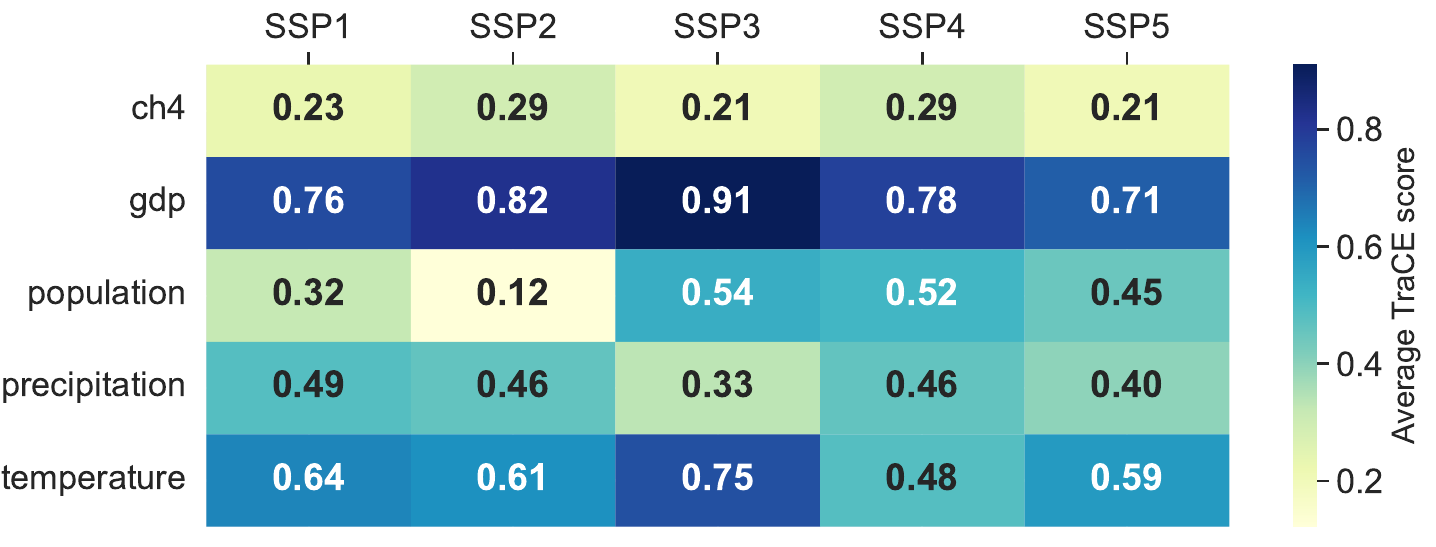}
    \caption{Feature heatmap of average TraCE scores for each SSP, in Poland, for the period 2015-2022.}
    \label{fig:appendix_trace_heatmap}
\end{figure}

\paragraph{SSP TraCE scores for Poland}
TraCE score analysis of the 34 countries in the global study found a common pattern across most countries, with SSP5 (Fossil-fueled Development) alignment ranking highest (Figure \ref{fig:ssp_trace_heatmap}). Several countries deviated from this pattern, such as Poland, for which the TraCE score time series is shown in Figure \ref{fig:appendix_trace_timeseries}. The TraCE score for SSP4 (Inequality) is consistently high throughout the time series, with SSP3 (Regional Rivalry) closely tracking, and overtaking in some instances. SSP4 then begins to diverge, leading as the highest ranked SSP from 2019 onwards. Unlike other countries in the study, SSP5 (Fossil-fueled Development) and SSP1 (Sustainability) are consistently ranked lowest throughout the time series. Interpretation of these results can be informed by Figure \ref{fig:appendix_trace_heatmap}, which shows the feature-level heatmap of average SSP TraCE scores over the study period (2015-2022). These scores have been determined by applying the TraCE method to each feature individually, to indicate their sole alignment with the corresponding SSP projections for that feature. Note that due to the way in which TraCE is formulated, these scores are not linearly disaggregated from the overall TraCE score for the country. In Figure \ref{fig:appendix_trace_heatmap}, the high TraCE score for SSP4 is dominated by the GDP feature, with other features also scoring highly for this SSP. SSP3 is dominated by the GDP and temperature features. The heatmap also shows that the features are most closely aligned with their SSP projections for GDP, temperature, and precipitation, with poor alignment for methane (CH4) projections across all SSPs. 

\end{document}